\documentclass[letterpaper, 10 pt, conference]{ieeeconf} 

\IEEEoverridecommandlockouts
\usepackage[utf8]{inputenc}
\usepackage{graphicx}
\usepackage{amsmath,wrapfig}
\usepackage{amssymb,amsfonts}
\usepackage{xcolor}
\usepackage{tikz}
\usepackage{subcaption}
\usepackage{cancel}
\usetikzlibrary{scopes}

\graphicspath{{figures/}}

\title{\LARGE \bf Nailed It: Autonomous Roofing with a Nailgun-Equipped Octocopter}
\author{Matthew Romano$^{a}$, Yuxin Chen$^b$, Owen Marshall$^a$, Ella Atkins$^{a,b}$%
\thanks{$^a$Authors are with the Robotics Institute at the University of Michigan {\tt\small\{mmroma, oamarsh, ematkins\}@umich.edu}}%
\thanks{$^b$Authors are with the Department of Aerospace Engineering at the University of Michigan {\tt\small\{chyuxin, ematkins\}@umich.edu}}%
}
\date{September 2019}

\begin{document}

\maketitle

\begin{abstract}
This paper presents the first demonstration of autonomous roofing with a multicopter. A DJI S1000 octocopter equipped with an off-the-shelf nailgun and an adjustable-slope roof mock-up were used. The nailgun was modified to allow triggering from the vehicle and tooltip compression feedback. A mount was designed to adjust the angle to match representative roof slopes. An open-source octocopter autopilot facilitated controller adaptation for the roofing application.  A state machine managed autonomous nailing sequences using smooth trajectories designed to apply prescribed contact forces for reliable nail deployment. Experimental results showed that the system is capable of nailing within a required  three centimeter gap on the shingle. Extensions to achieve a complete autonomous roofing system are discussed as future work.
\end{abstract}

\section{Introduction}

Unmanned aircraft systems (UAS) have been used for a variety of perception tasks including aerial photography, surveillance, and forest fire monitoring. More recently, physical manipulation applications have been explored such as grasping, perching, and tethered payload carriage  \cite{Huber_Kondak_Krieger_Sommer_Schwarzbach_Laiacker_Kossyk_Parusel_Haddadin_Albu_Schaffer_2013}\cite{Tsukagoshi_Watanabe_Hamada_Ashlih_Iizuka_2015}\cite{Geng_Langelaan_2019}. UAS can access sites that are  hazardous and difficult for people to reach. One example is installing a roof on a house.

\begin{figure}[t]
  \centering
    \includegraphics[width=\columnwidth,trim=192 25 255 50, clip]{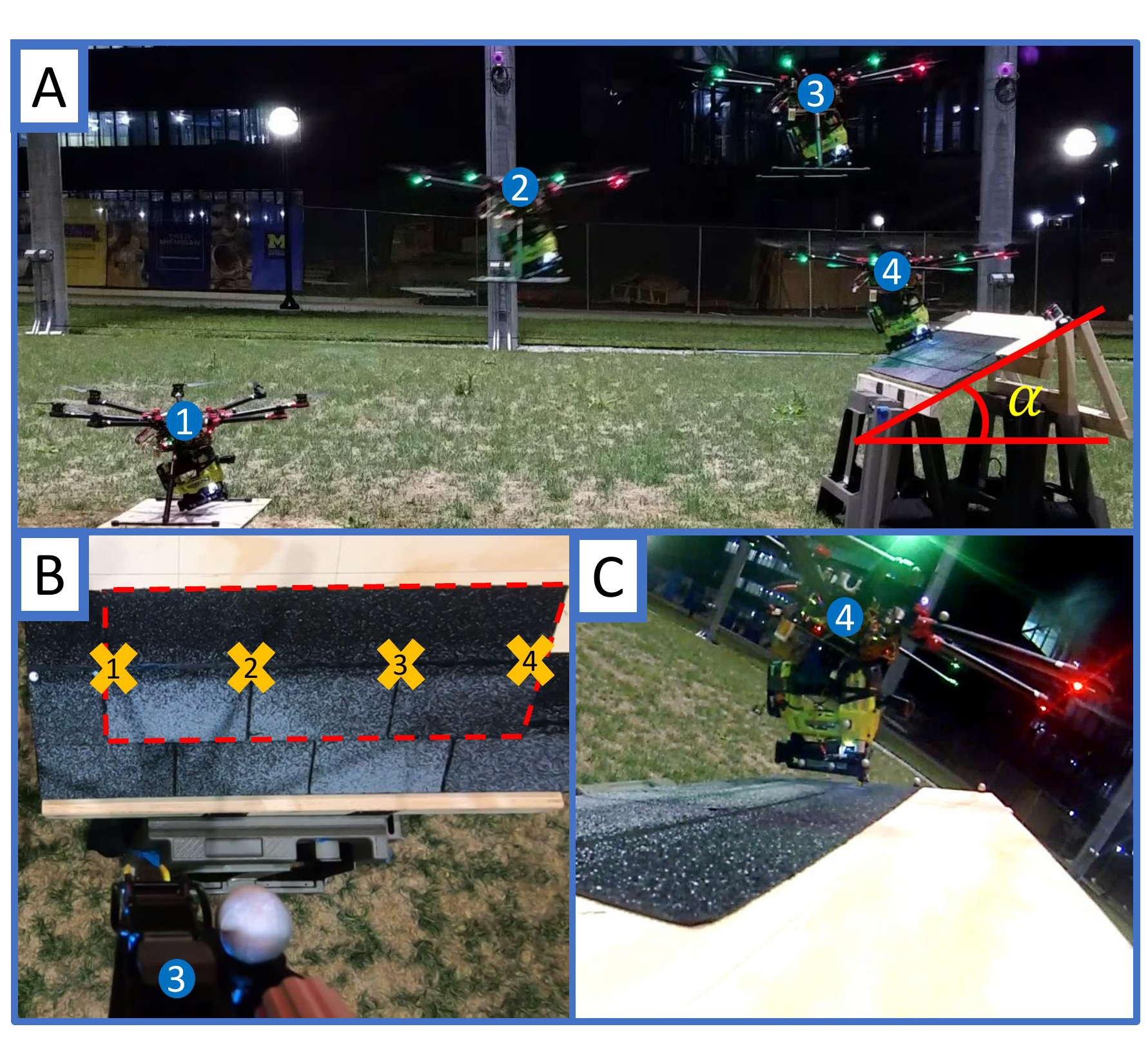}
  \caption{Autonomous roofing nailing sequence from three views. A) Third person view showing the vehicle at four time frames (denoted by the numbered blue circles). A roof mock-up is shown on the right  at an inclination angle of $\alpha$. B) Nailgun view during time frame 3. A placed shingle is ready to be nailed; the four nailing points are shown with yellow crosses. C) Roof board view during time frame 4. Contact has been made and nailing is about to occur.}
  \label{fig:main_figure}
\end{figure}

This paper presents the first autonomous experimental demonstration of roofing using a nailgun-equipped octocopter.
Specifically, we consider the scenario in which a human or robot assistant has already placed a shingle and the octocopter attaches the shingle to the roof. Figure \ref{fig:main_figure} shows the autonomous roofing sequence. 
The vehicle takes off per frame 1 and follows a prescribed trajectory through frame 2 to a hold waypoint at frame 3. The vehicle then must establish contact between the mounted nailgun and each of four nailing points as marked with a consistent force and sufficient accuracy to effectively fasten the shingle. The octocopter must also be capable of adjusting to different roof slopes while handling associated changes in applied forces. 

This paper offers the following contributions:
\begin{enumerate}
    \item Hardware integration for roof nailing via UAS.
    \item Characterization of UAS waypoints, motions, and contact forces required for successful nail insertion.
    \item First experimental demonstration of autonomous roofing with a UAS with error analysis.
\end{enumerate}

Section \ref{sec:RelatedWork} reviews recent work in multirotor guidance and control and aerial manipulation. Section \ref{sec:SystemDescription} presents our experimental system including a nailgun-equipped octocopter and a slope-adjustable roof mock-up. Section \ref{sec:Planner} describes our guidance algorithm and velocity nailing trajectory. Experimental results are shown in Section \ref{sec:Results}, while Sections \ref{sec:Discussion} and \ref{sec:Conclusion} analyze nailing accuracy and follow on work.

\section{Related Work}
\label{sec:RelatedWork}

Aerial robots are capable of following complex 3D trajectories with high accuracy and support sufficient payload sizes for a range of applications.
Ref. \cite{Mahony_Kumar_Corke_2012} provides a tutorial on modeling, estimation, and control of multirotor vehicles with focus on quadrotors. In \cite{Falanga_Mueggler_Faessler_Scaramuzza_2017}, Falanga et al. consider the aggressive flight of a quadrotor through a narrow gap using onboard sensing. They generate a trajectory that considers geometric, dynamic, and perception constraints and experimentally demonstrate their system with an 80\% success rate on gaps up to 45 degrees. In \cite{Mellinger_Kumar_2011}, Mellinger and Kumar formulate an optimization problem for minimum snap trajectory generation and control for quadrotors. Given a set of 3-D positions and yaw angles at specified times, a trajectory is generated that satisfies velocity, accleration, and input constraints while minimizing snap. A nonlinear controller is presented along with experimental results of a quadrotor flying through thrown and static hoops. 

More recently, aerial robots have been used for manipulation tasks. Unlike grounded robot arms, aerial manipulators do not have a fixed base and are subject to complex aerodynamic effects. 
\cite{Ruggiero_Lippiello_Ollero_2018} and \cite{Bonyan_Khamseh_Janabi_Sharifi_Abdessameud_2018} provide recent  literature surveys of the field.
In \cite{Vempati_Kamel_Stilinovic_Zhang_Reusser_Sa_Nieto_Siegwart_Beardsley_2018}, Vempati et al. present a spray-painting quadrotor. The system consists of a quadrotor with a spray gun on a pan-tilt mechanism. Power and paint are supplied via tethers. A full system is presented that involves modeling a 3D painting surface offline, generating the desired painted surfaces as well as robot commands, then actually applying paint to the surface. 
In \cite{Vlantis_Marantos_Bechlioulis_Kyriakopoulos_2015}, Vlantis et al. consider the problem of landing a quadrotor on an inclined platform of a moving ground vehicle. An MPC controller is used for experimental demonstration. Ref. \cite{Huber_Kondak_Krieger_Sommer_Schwarzbach_Laiacker_Kossyk_Parusel_Haddadin_Albu_Schaffer_2013} describes an aerial manipulation system consisting of a helicopter with an attached seven degree of freedom industrial robotic arm. Analysis of dynamic coupling between helicopter and arm was performed, and a control approach to counteract this coupling was demonstrated in an object grasping task.
\cite{Tsukagoshi_Watanabe_Hamada_Ashlih_Iizuka_2015} presents an aerial robot with door-opening capabilities with an initial perching approach to the door. Suction cups enable the vehicle to hang; a knob-twisting manipulator with appropriate rotor thrust are then coordinated to open the door.

This paper demonstrates autonomous roof shingle nailing with an octocopter UAS. This system is built on existing work in multicopter guidance and control. To the best of our knowledge the roofing application is new to this paper.

\section{System Description}
\label{sec:SystemDescription}
The main components of our autonomous roofing system are an octocopter, a nailgun, and a custom-built roof mock-up. Hardware components and necessary modifications are described in Section \ref{subsec:Hardware}, and a description of our software is provided in Section \ref{subsec:Software}. 

\subsection{Hardware}
\label{subsec:Hardware}

A DJI Spreading Wings S1000 octocopter with weight capacity sufficient to carry the nailgun was utilized for testing. The S1000 has eight 400kV 4114 Pro brushless DC motors each with 500W maximum power. 8kHz Electronic Speed Controllers (ESCs) drive the motors that turn pairs of 1552 foldable high strength plastic propellers \cite{DJI_Spec}. A 6S 10000mAh LiPo battery powers the vehicle, which can achieve 10min of flight time with the 9.25kg gross takeoff weight including nailgun assembly.

We chose the RYOBI 18-Volt ONE+ Lithium-Ion Cordless AirStrike 18-Gauge Brad Nailer as our testing nailgun.  
Professional roofing nailguns use larger, flat head nails with air hose connection required. Our tetherless system is a prototype that allowed us to focus on autonomous operation. In practice, the octocopter would have a professional nailgun with an air hose and power tether for extended endurance.

\begin{figure}
 	\centering
 	\begin{subfigure}[c]{0.48\columnwidth}
	    \includegraphics[width = \textwidth,trim=0 0 0 0, clip]{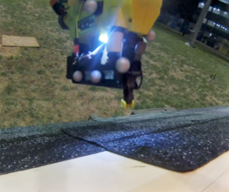}
	    \caption{Not Pressed, Light On} 
	    \label{fig:nail_gun_not_pressed_in}
    \end{subfigure}
  	\begin{subfigure}[c]{0.48\columnwidth}
	    \includegraphics[width = \textwidth,trim=0 0 0 0, clip]{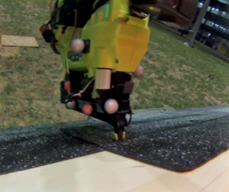}
	    \caption{Pressed, Light Off} 
	    \label{fig:nail_gun_pressed_in}
    \end{subfigure}
	\caption{Pressing the tooltip for nailing.} 
	\label{fig:nail_gun_pressing_in}
\end{figure}

The chosen nailgun came with built in safety features that require a specific procedure to deploy a nail: 1) Press the tooltip (Figure \ref{fig:nail_gun_pressing_in}), 2) Push the trigger button, and 3) Nail by keeping the tooltip pressed for about 0.5s. For autonomous operation, we followed the same procedure but with modification to step 2. A limit switch was added for feedback of tooltip state. The trigger was replaced with a relay controlled by our onboard electronics. While the vehicle follows its nailing trajectory, our software triggers the relay whenever the limit switch is pressed. At all other times, the relay does not trigger, disabling the nailgun and adding a layer of safety.

\begin{figure}
 	\centering
 	\begin{subfigure}[t]{0.48\columnwidth}
	    \includegraphics[width = \textwidth,trim=0 0 0 0, clip]{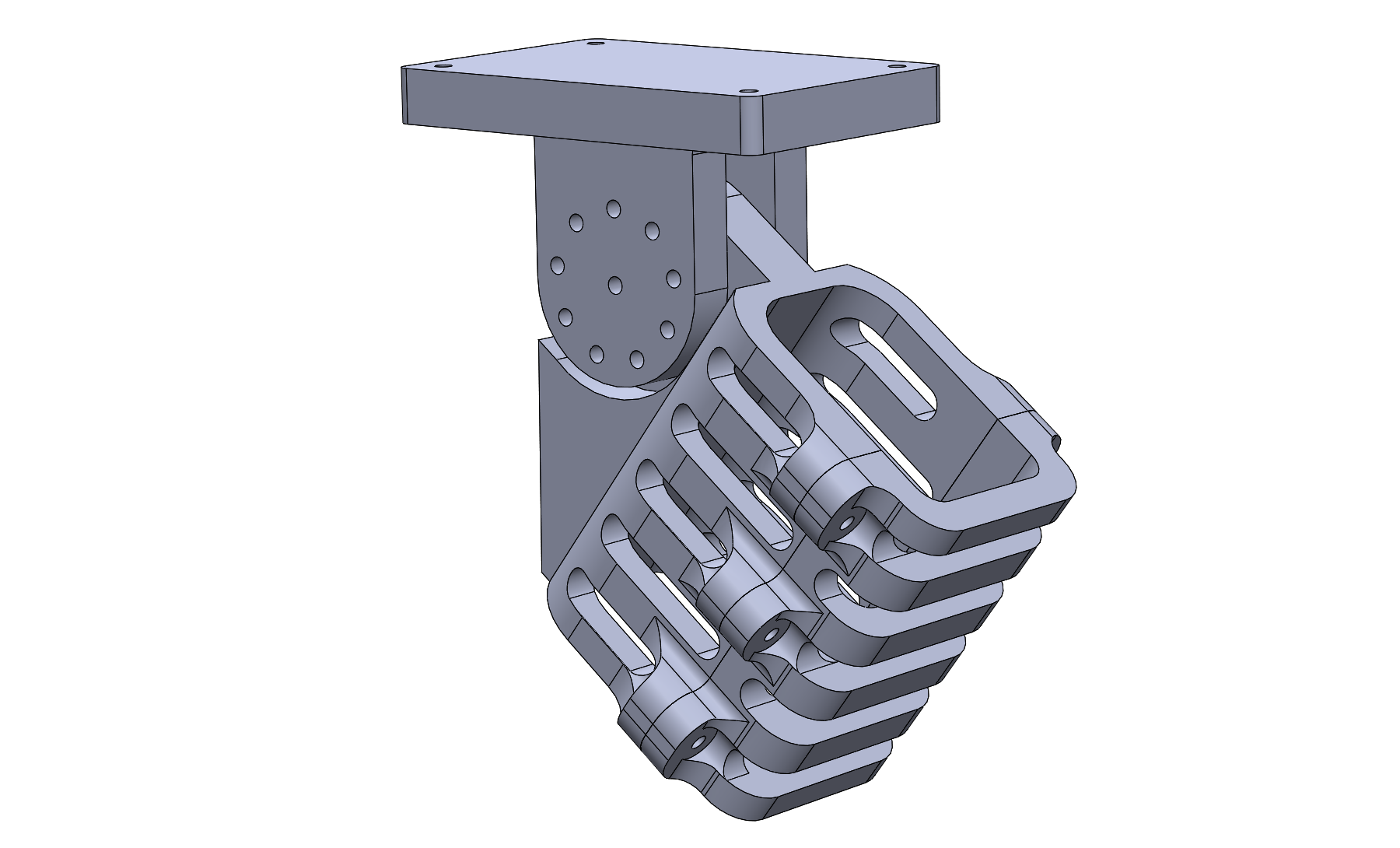}
	    \caption{Nailgun Mount} 
	    \label{fig:angles_are_fun_mount}
    \end{subfigure}
  	\begin{subfigure}[t]{0.48\columnwidth}
	    \includegraphics[width = \textwidth,trim=180 167 430 125, clip]{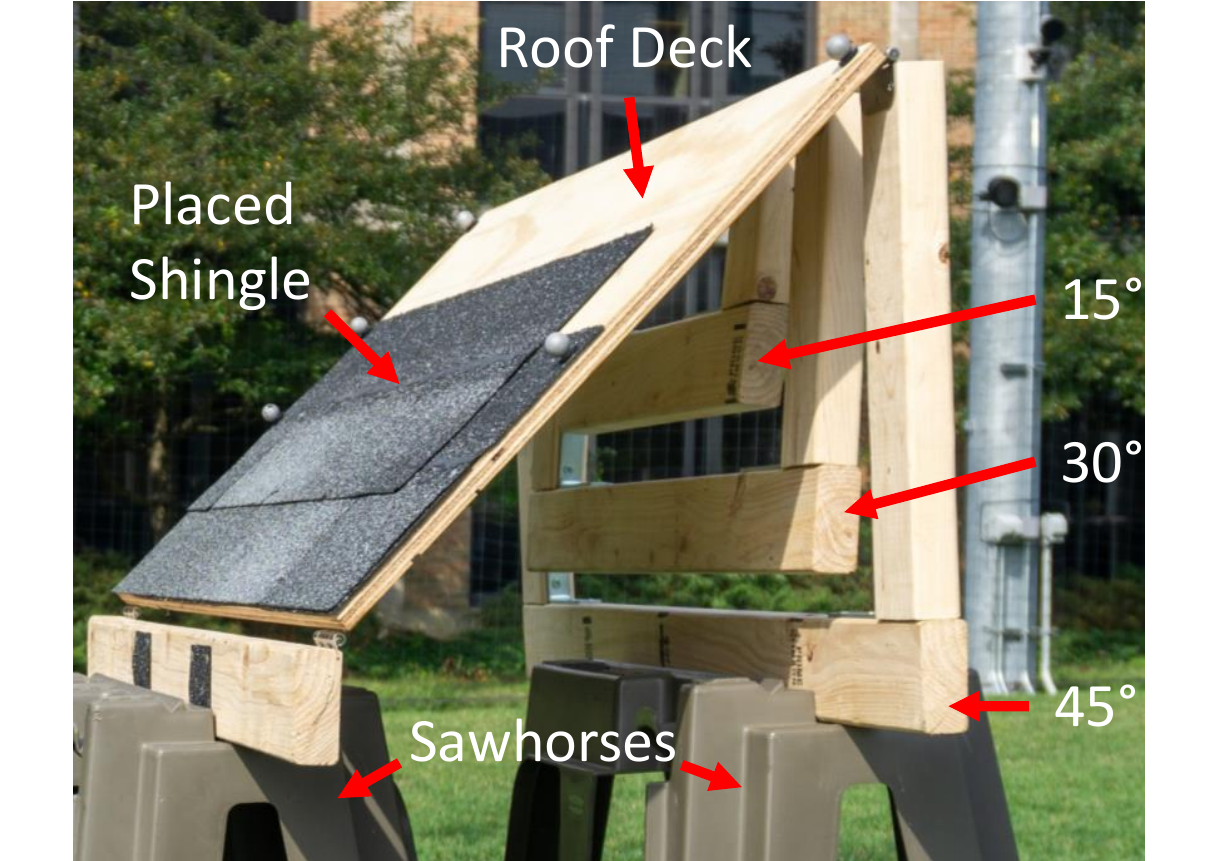}
	    \caption{Roof mock-up.} 
	    \label{fig:angles_are_fun_roof}
    \end{subfigure}
	\caption{Angle-adjustable experimental system. Both nailgun mount and roof mock-up can be set to 0$^\circ$, 15$^\circ$, 30$^\circ$, and 45$^\circ$.} 
	\label{fig:angles_are_fun}
\end{figure}

To demonstrate our system we constructed a roof mock-up per Figure \ref{fig:angles_are_fun_roof}. The roof deck is 3/4", 5-layer plywood with dimensions 2'$\times$4'. This decking is connected with hinges to three different stands with heights that support four different roof slope angles (0$^\circ$ (no stand), 15$^\circ$, 30$^\circ$, and 45$^\circ$). During experiments, the roof model is placed on two sawhorses with four sandbags to keep them stable. Half of the roof deck has attached shingles, and for each test a new shingle is placed but not yet nailed. The roof mock-up is carried flat.

Figure \ref{fig:angles_are_fun_mount} shows the nailgun mount. We designed a custom nailgun holder to mount the nailgun with adjustable angles corresponding to the available roof slopes (0$^\circ$ (straight down), 15$^\circ$, 30$^\circ$, and 45$^\circ$). All parts are 3D-printed.

\subsection{Software}
\label{subsec:Software}

We used a modified version of Ardupilot (APM) running on an embedded BeagleBone Blue single-board Linux computer. Its cascaded proportional-integral-derivative (PID) controller tracks a preloaded trajectory using motion capture feedback in M-Air, Michigan's outdoor netted motion capture equipped flight facility. See \cite{romano2019experimental} for further multicopter controller details.

\section{System Model}
\label{sec:SystemModel}

\subsubsection{Rigid Body Transformation} 
A rigid body transformation from frame $b$ to frame $a$ is defined as a rotation $\mathbf{R}_{b}^{a} \in \mathbb{R}^{3 \times 3}$ followed by a translation of $\mathbf{t}_{b}^{a} \in \mathbb{R}^{3 \times 1}$. A point $i$ expressed in frame $b$ ($\mathbf{r}_{i}^{b} \in \mathbb{R}^{3 \times 1}$) can be transformed to frame $a$ using $ \mathbf{r}_{i}^{a} = \mathbf{R}_{b}^{a} \mathbf{r}_{i}^{b} + \mathbf{t}_{b}^{a}$.

\subsubsection{Euler Angles} An Euler angle rotation ($\phi$,$\theta$,$\psi$) from frame $a$ to frame $b$ is a rotation $\psi$ (yaw) about the $z$-axis of frame $a$, followed by rotation  $\theta$ (pitch) about $y$, and finally rotation $\phi$ (roll) about $x$ resulting in 
\begin{align*}
    {\bf R}_{b}^{a} &={\bf R}_{z,\psi}{\bf R}_{y,\theta}{\bf R}_{x,\phi} \\
    &=
    \begin{bmatrix}
        c_\psi c_\theta & c_\psi s_\theta s_\phi-c_\phi s_\psi & s_\psi s_\phi + c_\psi c_\phi s_\theta\\
        c_\theta s_\psi & c_\psi c_\phi + s_\psi s_\theta s_\phi & c_\phi s_\psi s_\theta - c_\psi s_\phi\\
        -s_\theta & c_\theta s_\phi & c_\theta c_\phi
    \end{bmatrix}.
\end{align*}

\begin{figure}[t]
  \centering
    \includegraphics[width=\columnwidth,trim=140 140 230 70, clip]{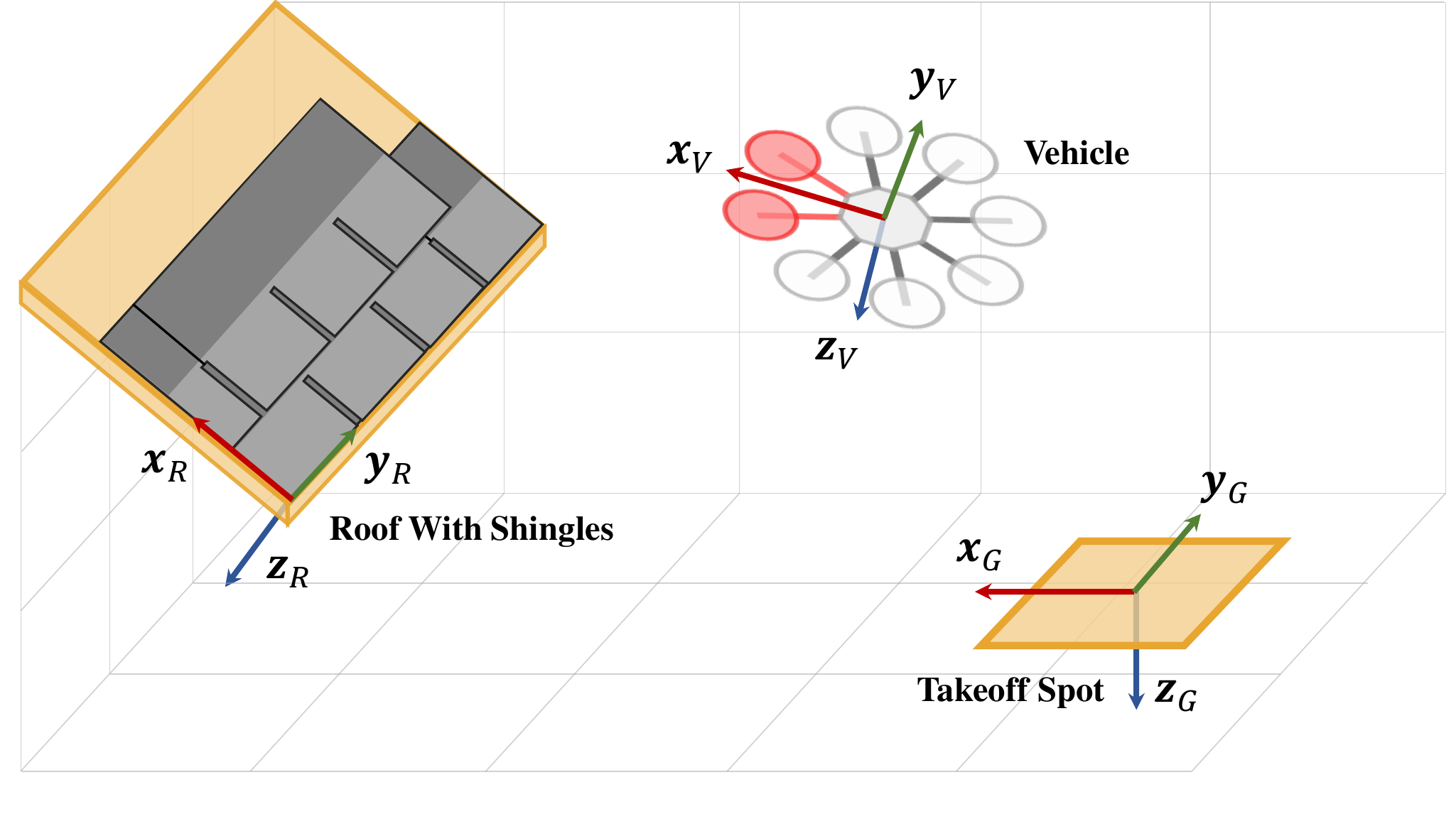}
  \caption{Reference Frames}
  \label{fig:FoR}
\end{figure}

Figure \ref{fig:FoR} shows relevant reference frames including the ground frame ($G$), roof frame ($R$), and vehicle frame ($V$). The ground frame follows a North-East-Down (NED) convention and is affixed to the takeoff position, which is also the motion capture system origin. 
The roof frame origin is located at the lower left corner of the roof which allows us to express the nailing points on the 2D roof surface with only $\mathbf{x}_{R}$ (vertical) and $\mathbf{y}_{R}$ (horizontal) values. The vehicle frame is attached to the center of the octocopter rotor plane.

Define transformations for the vehicle relative to the ground frame as  ($\mathbf{t}_{V}^{G}, \mathbf{R}_{V}^{G}=\mathbf{R}_{z,\psi_{V}} \mathbf{R}_{y,\theta_{V}} \mathbf{R}_{x,\phi_{V}}$) and roof relative to ground as ($\mathbf{t}_{R}^{G}, \mathbf{R}_{R}^{G}=\mathbf{R}_{z,\psi_{R}} \mathbf{R}_{y,\theta_{R}} \mathbf{R}_{x,\phi_{R}}$). Both are obtained from motion capture (with $\mathbf{R}_{V}^{G}$ also estimated from an onboard IMU). Define $\alpha = \theta_{R}$ per Figure \ref{fig:main_figure}.

\begin{figure}[htbp]
    \centering
    \includegraphics[width=\linewidth]{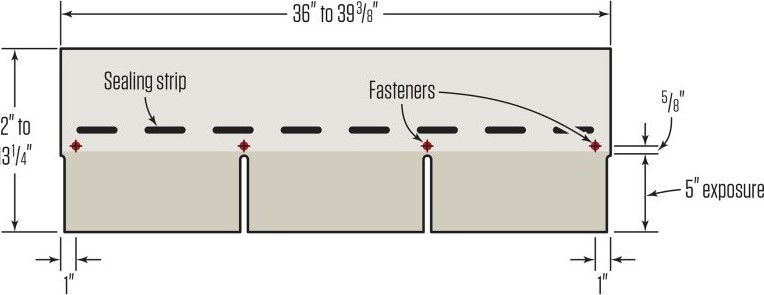}
    \caption{The Four-Nail Method for Three-Tab Shingles \cite{roofref1}.} 
    \label{fig:shingle}
\end{figure}

Our experiments used three-tab shingles and the four-nail method as seen in Figure \ref{fig:shingle} and explained in \cite{roofref1}. Nails should be placed on the shingle ideally just below the sealing strip and above the exposure cutouts (3cm gap). An optimal placement of the four nails is shown in Figure \ref{fig:shingle}.   

\begin{figure}[t]
  \centering
    \includegraphics[width=0.6\columnwidth,trim=325 125 325 125, clip]{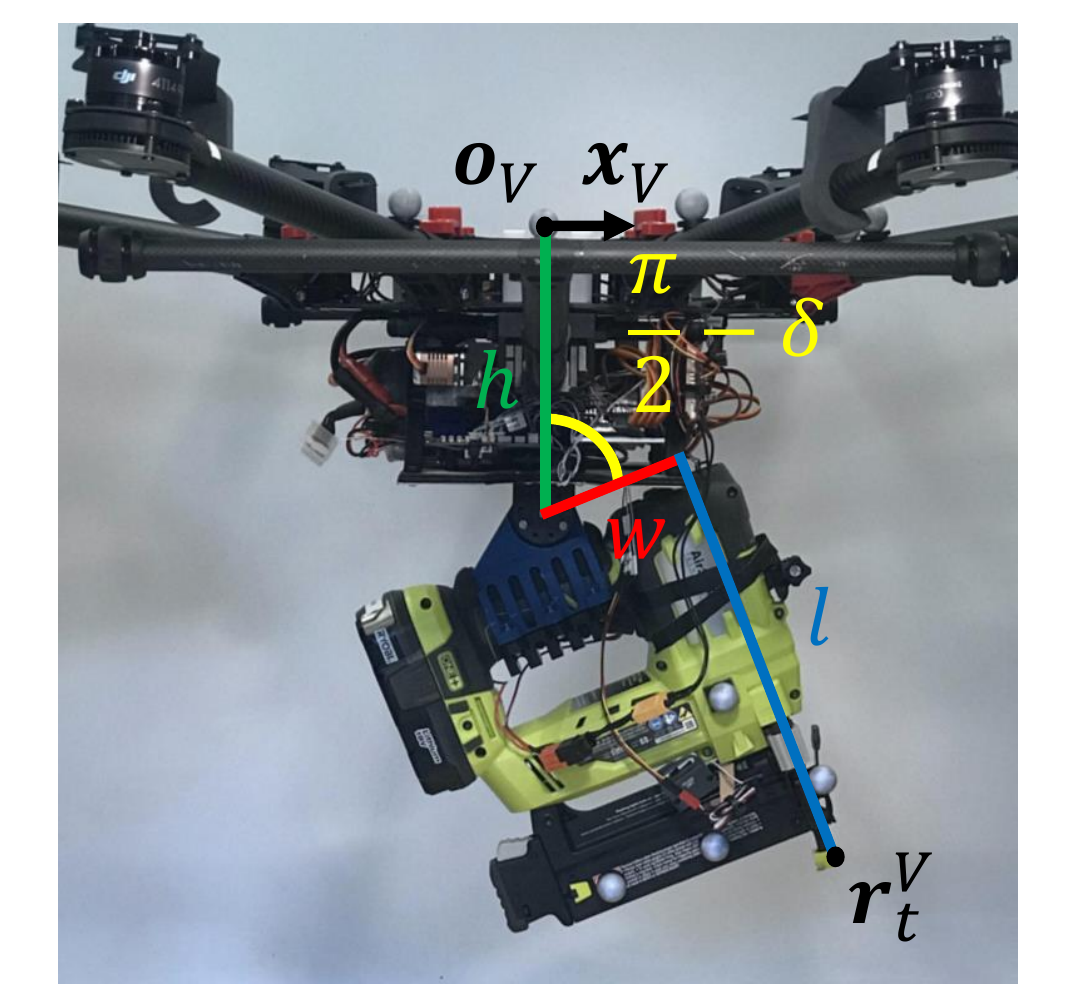}
  \caption{Tooltip position in vehicle frame.}
  \label{fig:Tip2BC}
\end{figure}

Define $N$ nailing points in the roof frame as

\begin{equation}
    {\bf r}^{R}_{n,i} = 
    \begin{bmatrix}
        x_{n,i}^{R} \\ 
        y_{n,i}^{R} \\
        0 
    \end{bmatrix}, 
    i \in \{1,...,N\},
\end{equation}

\noindent where $x_{n,i}^R$ and $y_{n,i}^R$ are the coordinates of the $i$th point in $\mathbf{x}_{R}$ and $\mathbf{y}_{R}$ respectively. These can be converted to the ground frame with $\mathbf{r}_{n,i}^{G} =  \mathbf{R}_{R}^{G} \mathbf{r}_{n,i}^{R}  + \mathbf{t}_{R}^{G}$.
From Figure \ref{fig:Tip2BC} we can describe tooltip position in the vehicle frame as

\begin{equation}
    \mathbf{r}_{t}^{V} =
    \begin{bmatrix}
        w \cos{\delta} + l \sin{\delta} \\
        0 \\
        h - w \sin{\delta} + l \cos{\delta} \\
    \end{bmatrix},
\end{equation}

\noindent which can also be converted into the ground frame using $\mathbf{r}_{t}^{G} =  \mathbf{R}_{V}^{G} \mathbf{r}_{t}^{V}  + \mathbf{t}_{V}^{G}$.
To place the tooltip to be at nailing point $i$, $\mathbf{r}_{t}^{G}$ must equal $\mathbf{r}_{n,i}^{G}$, so we have:
 
\begin{equation*}
     \mathbf{R}_{V}^{G} \mathbf{r}_{t}^{V}  + \mathbf{t}_{V}^{G} = \mathbf{R}_{R}^{G} \mathbf{r}_{n,i}^{R}  + \mathbf{t}_{R}^{G} \\ 
\end{equation*}

\vspace{0cm}

\begin{equation}
    \mathbf{t}_{V}^{G} = \mathbf{R}_{R}^{G} \mathbf{r}_{n,i}^{R}  + \mathbf{t}_{R}^{G} - \mathbf{R}_{V}^{G} \mathbf{r}_{t}^{V} 
    \label{eq:vehiclePosition4Nailing},
\end{equation}

\noindent which defines position of the vehicle in the ground frame for each nailing location $i$.

\section{Guidance}
\label{sec:Planner}

\subsubsection{Quintic Spline Method}
Consider a fifth order polynomial of the form
\begin{equation}
    r_i(t)=\sum\limits_{j=0}^5 a_{j,i}t^j,
\end{equation}
where $r_i(t)$ is the trajectory function at travel time $t$ in the $i$th direction. To solve for the six coefficients, boundary conditions at two times, $t_0$ and $t_f$ can be provided. The coefficients can then be obtained by solving
\begin{equation}
    \begin{bmatrix}
    1&t_i&t_0^2&t_0^3&t_0^4&t_0^5\\
    0&1&2t_0&3t_0^2&4t_0^3&5t_0^4\\
    0&0&2&6t_0&12t_0^2&20t_0^3\\
    1&t_f&t_f^2&t_f^3&t_f^4&t_f^5\\
    0&1&2t_f&3t_f^2&4t_f^3&5t_f^4\\
    0&0&2&6t_f&12t_f^2&20t_f^3\\
    \end{bmatrix}\begin{bmatrix}
    a_{0,i}\\a_{1,i}\\a_{2,i}\\a_{3,i}\\a_{4,i}\\a_{5,i}
    \end{bmatrix}=\begin{bmatrix}
    r_i(t_0)\\\dot{r_i}(t_0)\\\ddot{r_i}(t_0)\\r_i(t_f)\\\dot{r_i}(t_f)\\\ddot{r_i}(t_f)
    \end{bmatrix}.
\end{equation}

\subsubsection{Constant Velocity Trajectory from Rest}
We designed a straight line, constant velocity trajectory to move the vehicle from rest to a specified velocity ($v_f$) while respecting maximum acceleration constraints ($a_{max}$) through a distance $\Delta s$. This is first done in 1D and then projected into 3D. We consider a trajectory of the form

\begin{equation}
    s_{1}(t) = b_{3} t^{3} + b_{4} t^{4}. 
\end{equation}

\noindent The extrema of $\ddot{s}_{1}(t)$ occur at $t_{ext} = -\frac{b_{3}}{4b_{4}}$. We constrain $\ddot{s}_{1}(t_{ext}) = a_{max}$, $\ddot{s}_{1}(t_{1}) = 0$, and $\dot{s}_{1}(t_{1 }) = v_f$ then solve for the parameters and the required time to reach desired velocity ($t_{1}$) as

\begin{equation}
    b_{3} = \frac{4a_{max}^2}{9v_{f}} \text{, } 
    b_{4} = -\frac{4a_{max}^3}{27 v_{f}^2} \text{, }
    t_{1} = \frac{3v_{f}}{2a_{max}}.
\end{equation}

\noindent The remaining trajectory has constant velocity until $t_{f}$, where

\begin{equation}
    t_{f} = t_1 + \frac{\Delta s - s_{1}(t_1)}{v_f}. \\ 
\end{equation}

\noindent The full trajectory in 1D is

\begin{equation}
    s(t) = 
    \begin{cases}
        b_{3} t^{3} + b_{4} t^{4} & 0 \leq t \leq t_1 \\
        s_{1}(t_1) +  v_{f} \left( t - t_1 \right) & t_1 \leq t \leq t_{f},  \\
    \end{cases}
\end{equation}

\noindent and can be projected into 3D from $\mathbf{r}_a$ to $\mathbf{r}_b$ as

\begin{equation}
    \mathbf{r}(t) = \frac{\mathbf{r}_b - \mathbf{r}_a }{||\mathbf{r}_b - \mathbf{r}_a || } s(t) + \mathbf{r}_{a} \text{, } 0 \leq t \leq t_{f}
    \label{eq:constant_vel_traj}.
\end{equation}

\subsubsection{Desired Nailing Velocity}
We assumed conservation of energy during tooltip compression to obtain the desired velocity magnitude for each boundary condition.
\begin{equation}
    \frac{1}{2}m v_1^2 = \cancelto{0}{\frac{1}{2}mv_2^2} +  
    \frac{1}{2}kc^2\Rightarrow v_{1} = \sqrt{\frac{k}{m}}c,
    \label{eq:solve_for_vel}
\end{equation}
where $m$ is the total system mass, $v_1$ is the velocity before compression, $v_2$ is the final velocity after compression (zero), $k$ is the spring constant, and $c$ is the compression length.
The tooltip spring constant was measured by vertically pushing down the nailgun on a scale until it compressed. Compression distance was measured with a caliper. The remaining parameters are listed in Table \ref{tab:nailing_vel_calc} with a resulting desired nailing velocity of $13.6$cm/s. We chose $15$cm/s for experiments to ensure sufficient compression for nail release.
\begin{table}[htbp]
    \centering
    \caption{Tooltip and Nailing Velocity Parameters}
    \begin{tabular}[pos]{c | c | c}
     Notation & Name & Value \\\hline
     $F$ & Spring Force & 25N \\
     $c$ & Compression Distance & 7mm \\
     $k$ & Spring Const. & 3.5kN/m \\
     $m_o$ & Octocopter Mass& 4.8kg \\
     $m_b$ & Battery Mass & 1.4kg \\
     $m_n$ & Nailgun Mass & 3.0kg \\
     $m$ & Total Mass & 9.2kg \\
     $v_1$ & Calculated Velocity & 13.6cm/s
    \end{tabular}
    \label{tab:nailing_vel_calc}
\end{table}

\subsubsection{Nailing Trajectory}
\begin{figure}[t]
  \centering
    \includegraphics[width=\columnwidth,trim=145 90 200 93, clip]{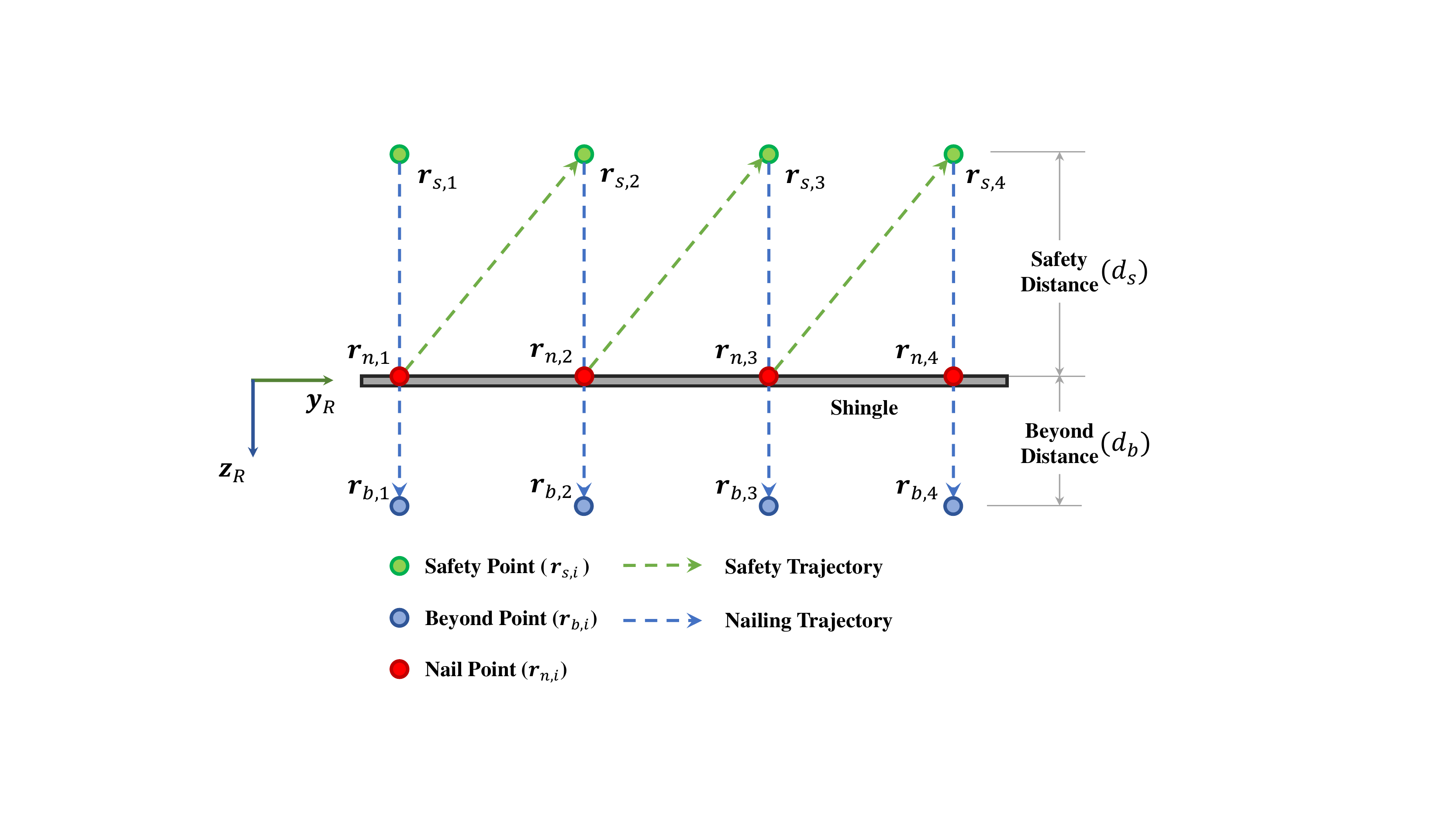}
  \caption{Nailing trajectory in the roof frame.}
  \label{fig:trajStage}
\end{figure}

Successful nail deployment requires the UAS to keep the tooltip at the nailing point while maintaining sufficient compression force until it fires. We achieved this goal with a tooltip trajectory for each nail point that starts at a safety point, $\mathbf{r}_{s,i}$, goes through the desired nailing point, $\mathbf{r}_{n,i}$, and ends at a "beyond point" $\mathbf{r}_{b,i}$. Figure \ref{fig:trajStage} shows our method. The tooltip doesn't actually reach $\mathbf{r}_{b,i}$, but this position error allows for the cascaded PID controller to apply a force in a controlled manner. $\mathbf{r}_{s,i}$ and $\mathbf{r}_{b,i}$ are obtained by projecting $\mathbf{r}_{n,i}$ up by $d_{s}$ and down by $d_{b}$ in the roof frame, respectively as

\begin{equation}
    \mathbf{r}^{R}_{s,i} = \mathbf{r}^{R}_{n,i} - 
    \begin{bmatrix}
        0 \\
        0 \\
        d_{s} \\
    \end{bmatrix}
    \text{, }
    \mathbf{r}^{R}_{b,i} = \mathbf{r}^{R}_{n,i} + 
    \begin{bmatrix}
        0 \\
        0 \\
        d_{b} \\
    \end{bmatrix}.
\end{equation}

\begin{figure}[t]
  \centering
    \includegraphics[width=0.7\columnwidth,trim=102 230 120 250, clip]{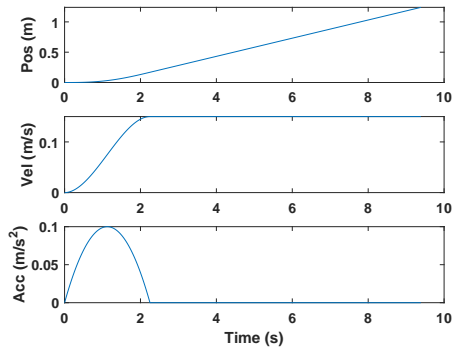}
  \caption{1D nailing trajectory consisting of a smooth acceleration up to a constant velocity held for nailing.}
  \label{fig:1d_traj}
\end{figure}

\noindent The constant velocity trajectory in Eq. \ref{eq:constant_vel_traj} was used for the nailing trajectory from $\mathbf{r}_{s,i}$ to $\mathbf{r}_{b,i}$ with $v_{f}=15$cm/s, $a_{max}=10$cm/s$^2$, $d_{s}=1$m, $d_{b}=24$cm and the 1D trajectory can be seen in Figure \ref{fig:1d_traj}. Quintic splines were used for the safety trajectory from $\mathbf{r}_{b,i}$ to $\mathbf{r}_{s,i+1}$ ($\mathbf{r}_{b,4}$ goes to $\mathbf{r}_{s,4}$) with zero velocity and acceleration at the endpoints. Lastly, a takeoff trajectory (takeoff point to $\mathbf{r}_{s,1}$), and a return and land trajectory ($\mathbf{r}_{s,4}$ to takeoff point) also used quintic splines for guidance. Eq. \ref{eq:vehiclePosition4Nailing} was used to convert the tooltip trajectory into a vehicle trajectory assuming hover conditions at the nailing point with $\phi_{V}=0$, $\theta_{V}=-2^\circ$, $\psi_{V}=\psi_{R}$, and $\delta=\alpha$. Vehicle hover pitch angle $\theta_{V}$ was obtained experimentally and is most likely non-zero due to the weight imbalance associated with the nailgun. 

\section{Results}
\label{sec:Results}

A total of four $\alpha$ angles were attempted to be nailed (0$^\circ$, 15$^\circ$, 30$^\circ$, and 45$^\circ$). However, the 45$^\circ$ case was too challenging and there were no successful nailing attempts. This is partly due to the tooltip slipping because there is less friction and partly due to the propellers getting too close to the board when nailing. For the three remaining cases a total of 16 experiments were performed with measured data over the course of several flight test days and a breakdown of the conditions can be seen in Table \ref{tab:results_conditions_breakdown} along with nail deployments in Table \ref{tab:nail_deployments}. For each experiment, four nailing attempts were made giving a total of 64 nailing attempts. Test set A was the most recent and most successful thus will be the focus of our discussion.

\begin{table}
    \centering
    \caption{Test set conditions.}
    \begin{tabular}[pos]{r | c | c | c | c | c}
     Test Set & Windy & PID Gains & 0$^\circ$ & 15$^\circ$ &  30$^\circ$ \\ \hline
     \textbf{A} & \textbf{No} &\textbf{set2} & \textbf{2} & \textbf{2} & \textbf{2} \\
     B & No & set2 &  0 & 5 & 0 \\
     C & Yes & set2 &  1 & 1 & 0 \\
     D & No & set2 &  0 & 0 & 1 \\
     E & No & set1 &  1 & 1 & 0 \\
    \end{tabular}
    \label{tab:results_conditions_breakdown}
\end{table}

\begin{table}
    \centering
    \caption{Test set nail deployments.}  
    \begin{tabular}[pos]{r | c | c }
     Test Set &  Nails Deployed & Nails Attempted \\ \hline
     \textbf{A} & \textbf{24} & \textbf{24} \\
     B & 17 & 20 \\
     C & 8  & 8 \\
     D & 4  & 4 \\
     E & 8  & 8 \\
    \end{tabular}
    \label{tab:nail_deployments}
\end{table}

Define horizontal error as $e_{h}= y_{d}^{R} - y_{a}^{R}$ and vertical error as $e_{v}= x_{d}^{R} - x_{a}^{R}$, where $x_{d}^{R} \text{ and } y_{d}^{R}$ are the desired nailing positions and $x_{a}^{R} \text{ and } y_{a}^{R}$ are where the nails actually were placed. Note that the desired nailing positions were shifted for various tests to attempt to correct for systematic bias, which may not align with the actual locations where the shingles should be nailed. Therefore, this definition of error allows us to compare all tests together even with the shifts.  After each experiment, the error of each nail was measured with respect to a shingle outline on the backside of the roof with marked setpoints. 

\begin{figure}[htb]
 	\centering
 	\begin{subfigure}[c]{0.48\columnwidth}
	    \includegraphics[width = \textwidth,trim=100 240 130 240, clip]{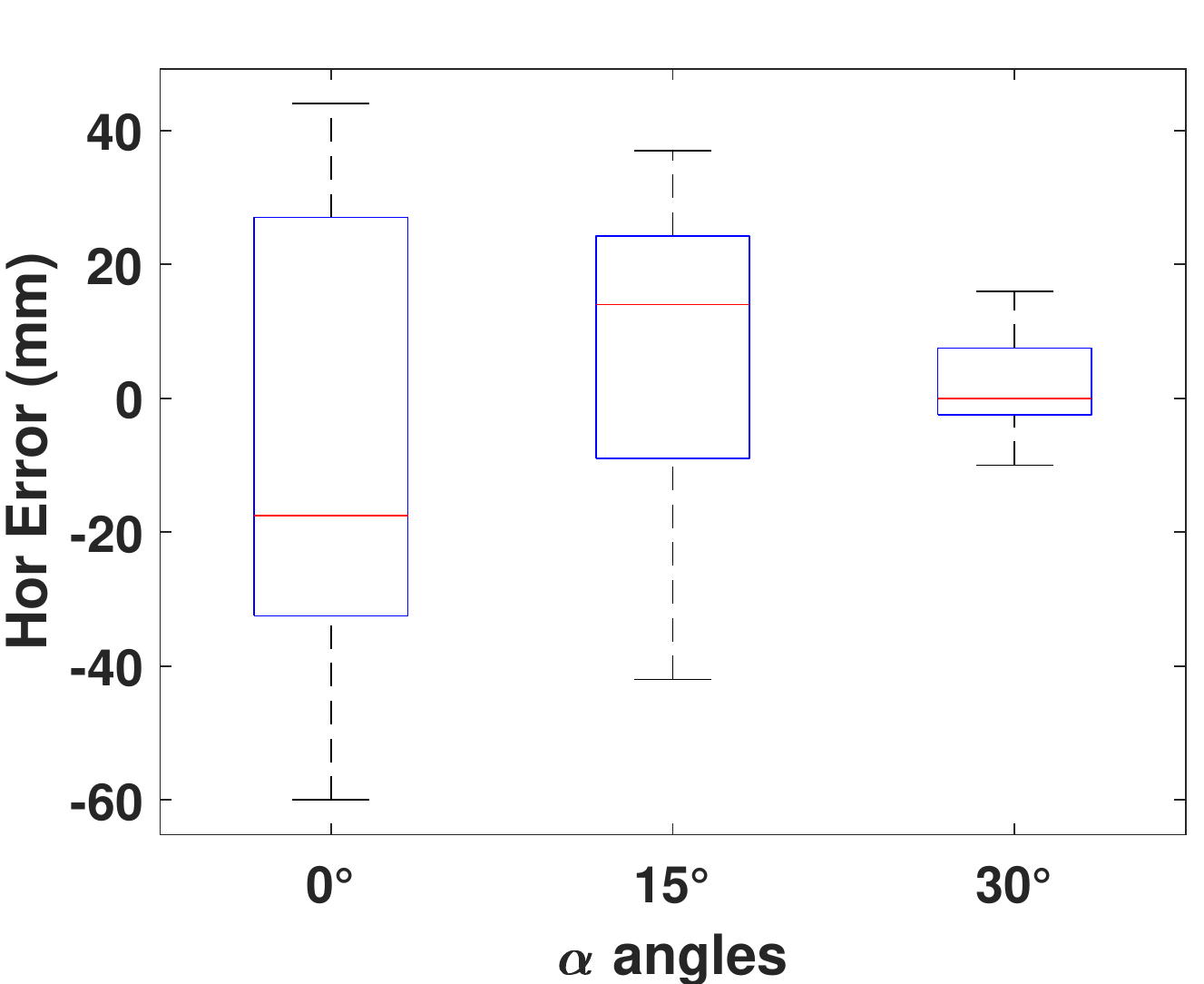}
	    \caption{Horizontal Error} 
	    \label{fig:all_tests_box_plot_hor_err}
    \end{subfigure}
  	\begin{subfigure}[c]{0.48\columnwidth}
	    \includegraphics[width = \textwidth,trim=100 240 130 240, clip]{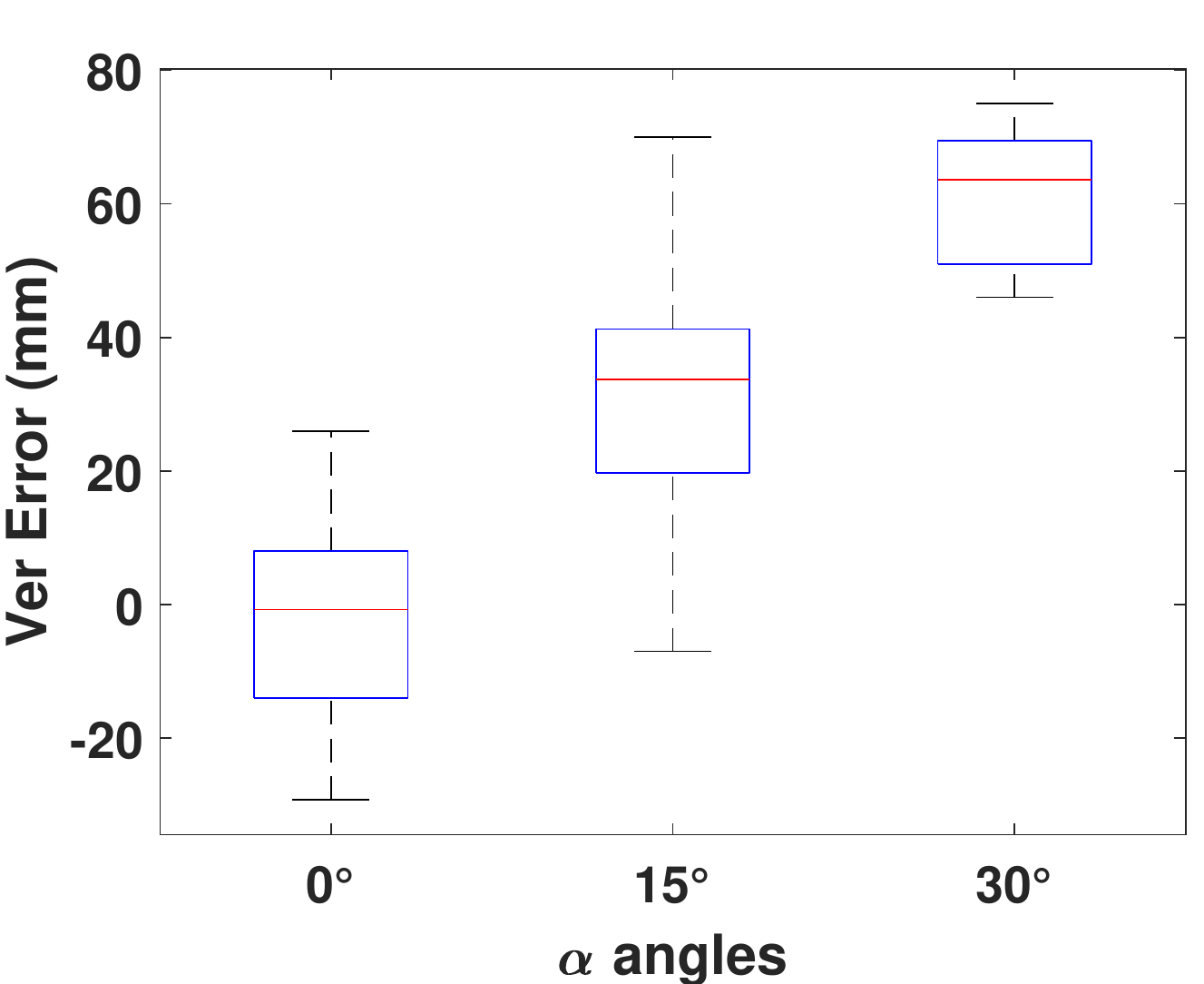}
	    \caption{Vertical Error} 
	    \label{fig:all_tests_box_plot_ver_err}
    \end{subfigure}
	\caption{All Tests: Nail placement error box plots by angle.} 
	\label{fig:all_tests_box_plot}
\end{figure}

Figure \ref{fig:all_tests_box_plot} shows the nailing accuracy for all tests. There is a trend in the horizontal error where the range decreases as the angle increases. For 0$^\circ$ there is low precision with a range of 10.4cm. For 30$^\circ$ the precision is high with a range of 2.6cm and the accuracy is also high with a median of 0.0cm. In vertical error there is a trend in accuracy. As the angle increases so does the median (-0.8cm at 0$^\circ$, 3.4cm at 15$^\circ$, and 6.4cm at 30$^\circ$). 

\begin{figure}[htb]
 	\centering
 	\begin{subfigure}[c]{0.48\columnwidth}
	    \includegraphics[width = \textwidth,trim=100 240 130 240, clip]{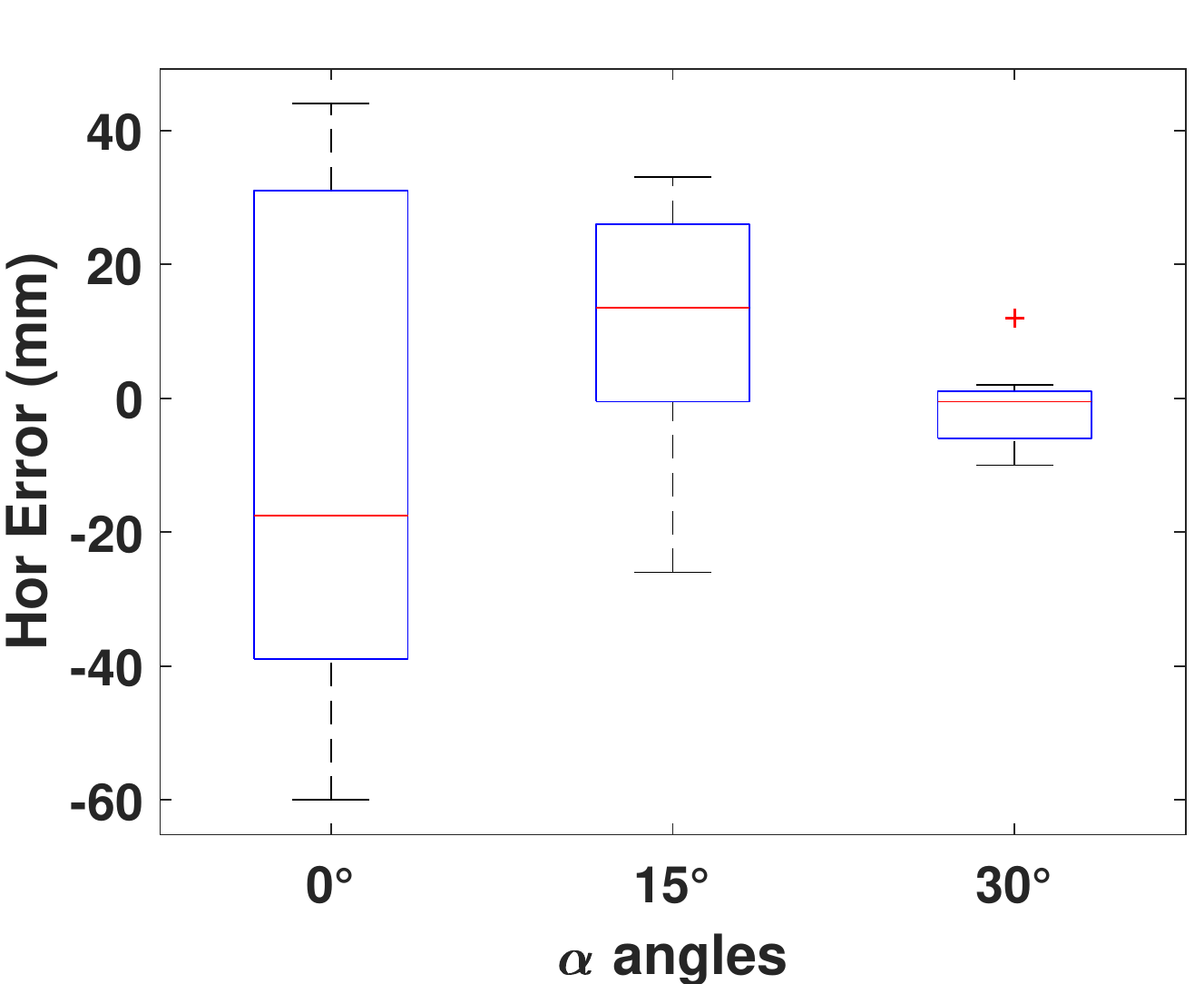}
	    \caption{Horizontal Error} 
	    \label{fig:testA_box_plot_hor_err}
    \end{subfigure}
  	\begin{subfigure}[c]{0.48\columnwidth}
	    \includegraphics[width = \textwidth,trim=100 240 130 240, clip]{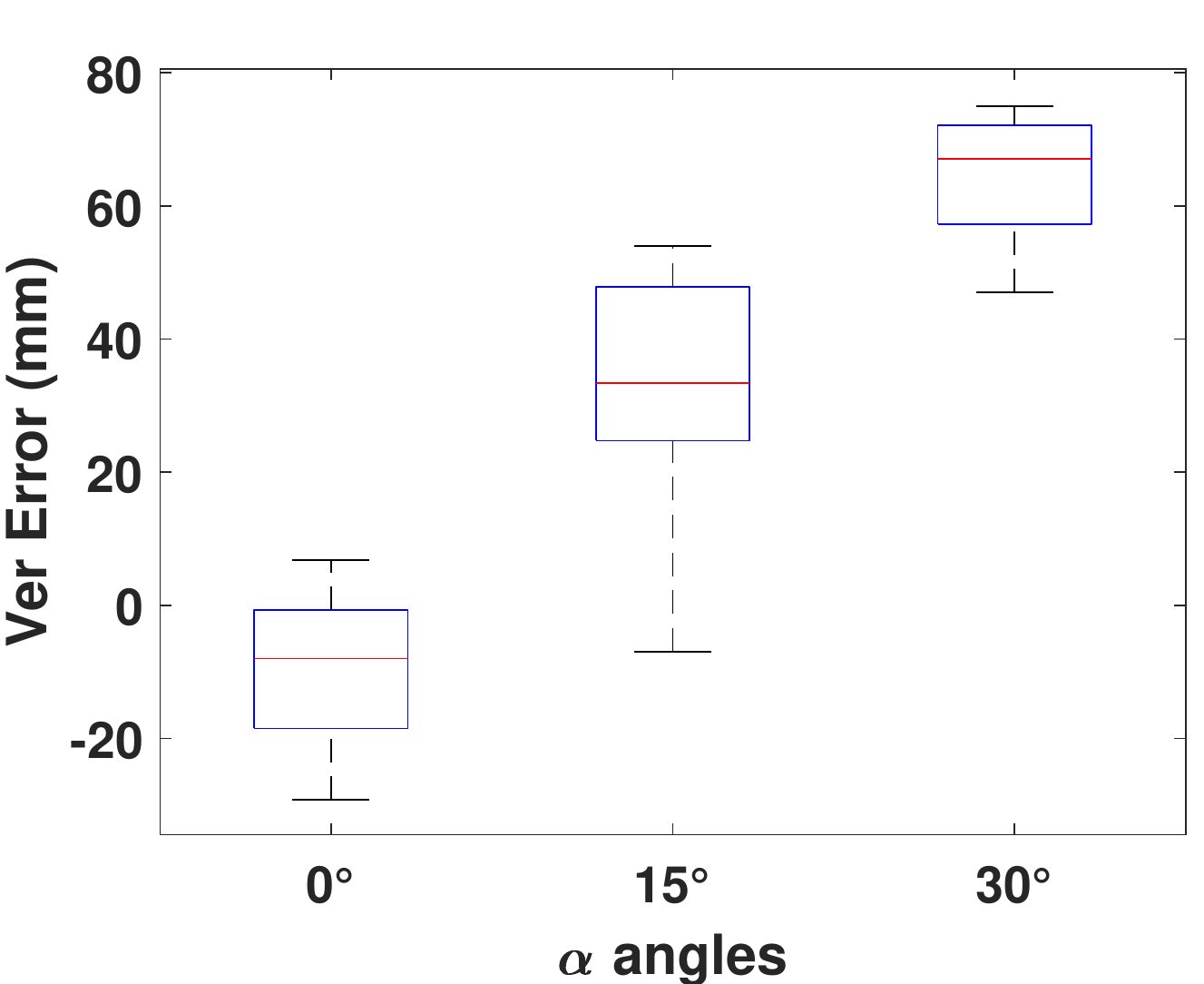}
	    \caption{Vertical Error} 
	    \label{fig:testA_box_plot_ver_err}
    \end{subfigure}
	\caption{Test Set A: Nail placement error box plots by angle.} 
	\label{fig:testA_box_plot}
\end{figure}

\begin{figure}[htb]
 	\centering
 	\begin{subfigure}[c]{0.9\columnwidth}
     	\centering
	    \includegraphics[width = 0.9\columnwidth,trim=100 270 100 300, clip]{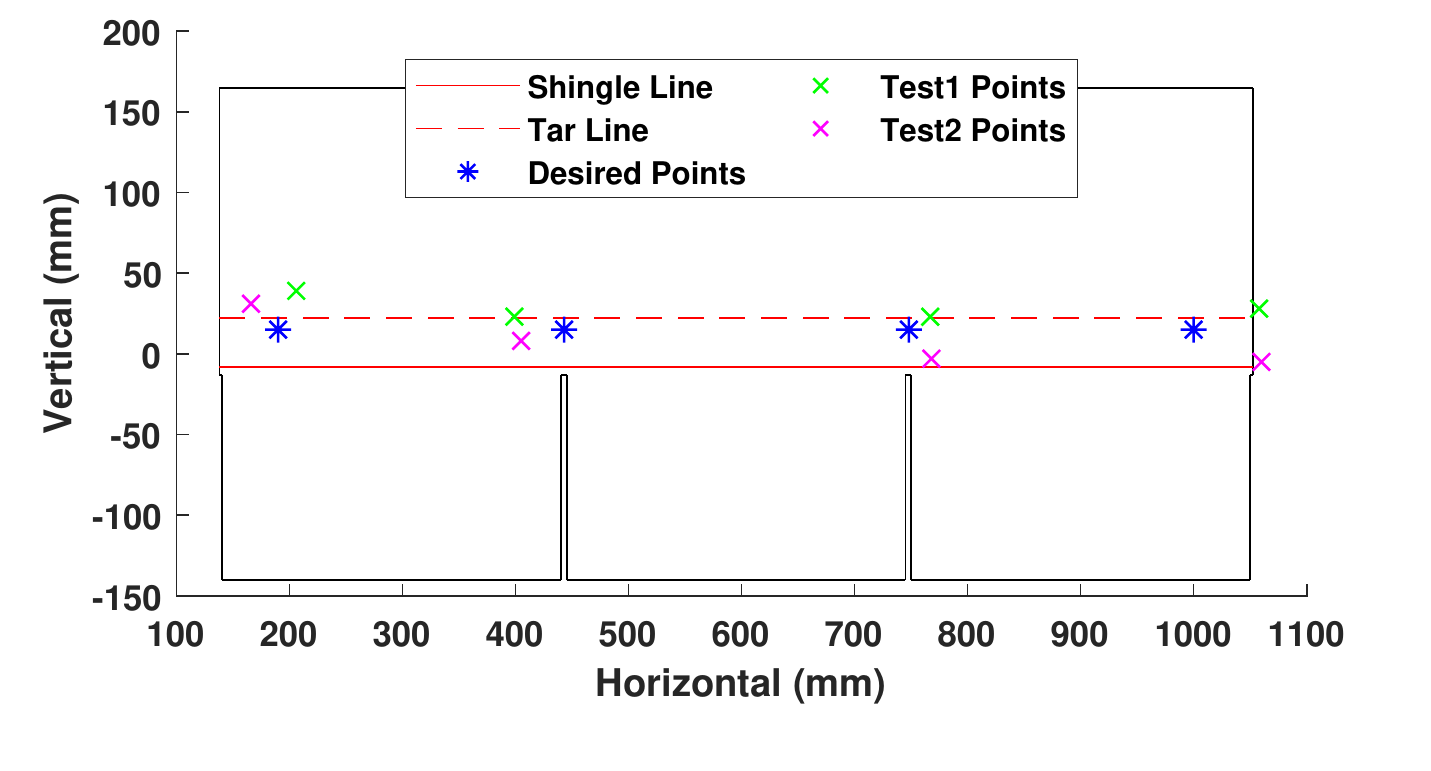}
	    \vspace{-0.5cm}
	    \caption{0 degrees} 
	    \label{fig:shingle_plot_setA_0}
    \end{subfigure}
    \vspace{0cm}
 	\begin{subfigure}[c]{0.9\columnwidth}
     	\centering
	    \includegraphics[width = 0.9\columnwidth,trim=100 270 100 300, clip]{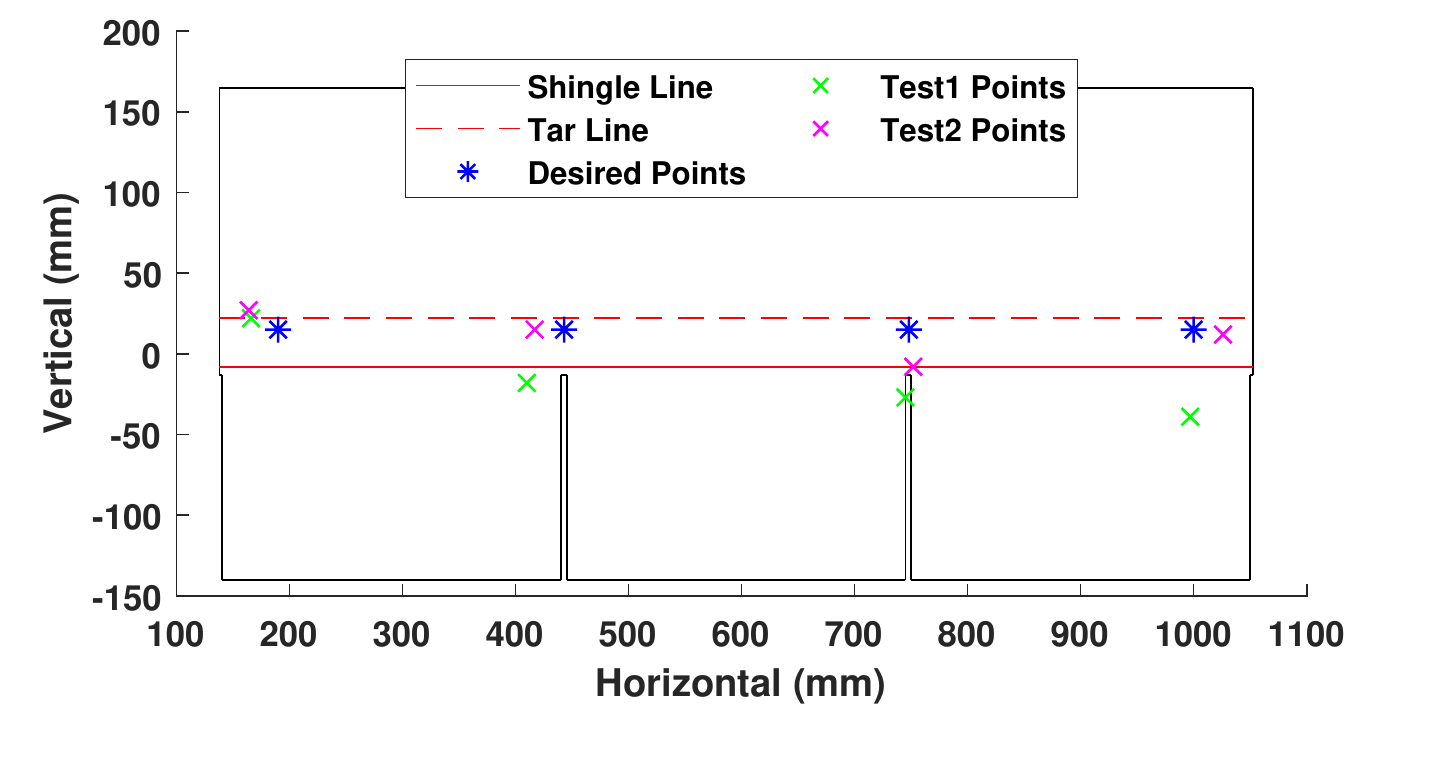}
	    \vspace{-0.5cm}
	    \caption{15 degrees} 
	    \label{fig:shingle_plot_setA_15}
    \end{subfigure}
    \vspace{0cm}
 	\begin{subfigure}[c]{0.9\columnwidth}
     	\centering
	    \includegraphics[width = 0.9\columnwidth,trim=100 270 100 300, clip]{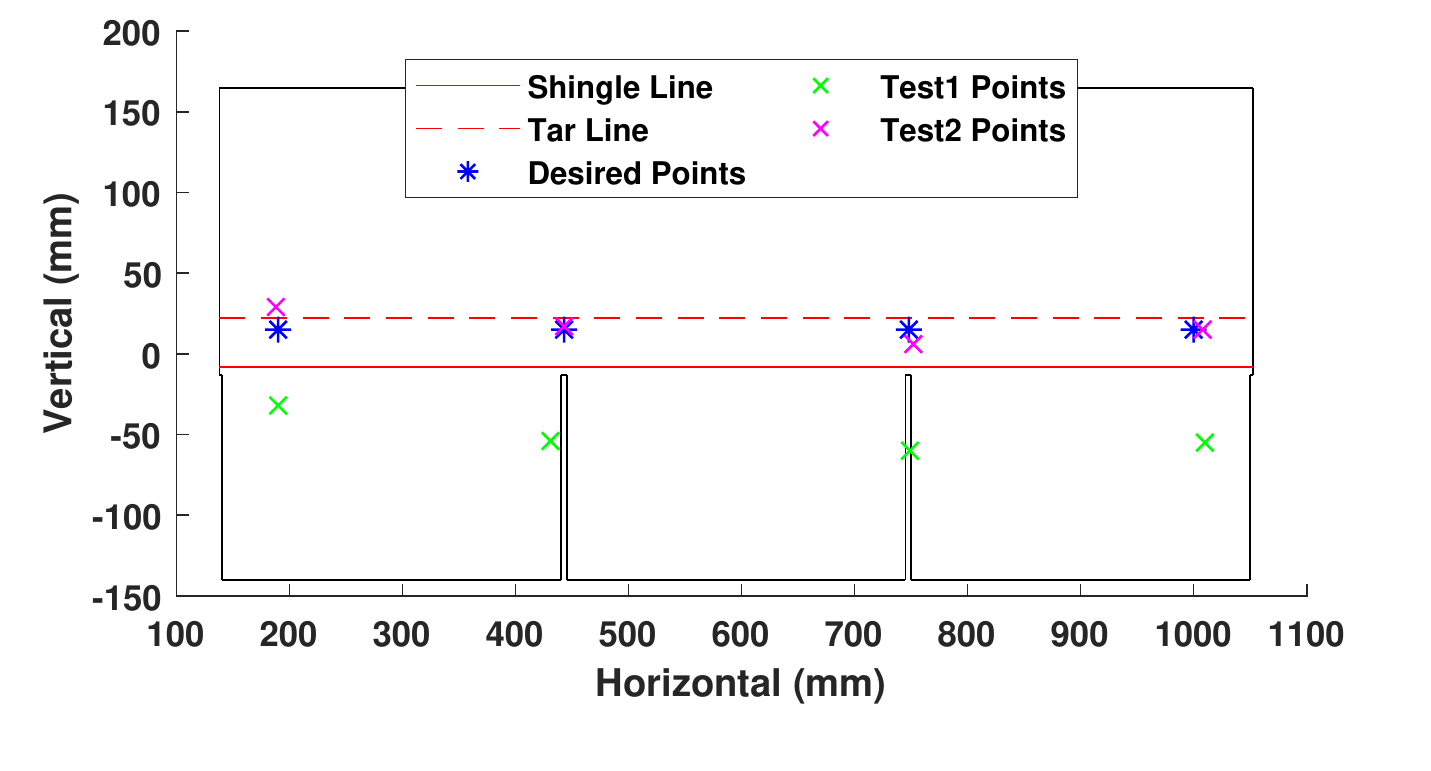}
	    \vspace{-0.5cm}
	    \caption{30 degrees} 
	    \label{fig:shingle_plot_setA_30}
    \end{subfigure}
	\caption{Nail placements for test set A at each angle.} 
	\label{fig:shingle_plot_setA}
\end{figure}

\begin{figure}[htb]
  \centering
    \includegraphics[width=0.75\columnwidth,trim=25 185 44 185, clip]{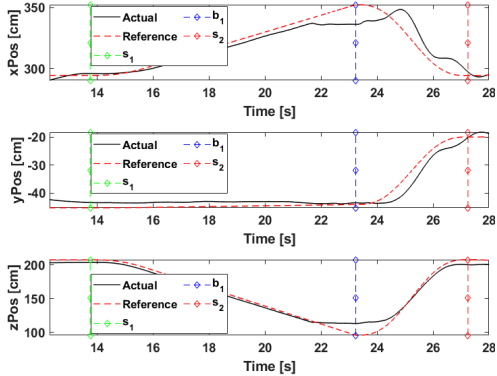}
  \caption{Actual Nailing Trajectory - Position Plot.}
  \label{fig:actual_traj_pos}
\end{figure}

\begin{figure}[htb]
  \centering
    \includegraphics[width=0.75\columnwidth,trim=30 185 44 185, clip]{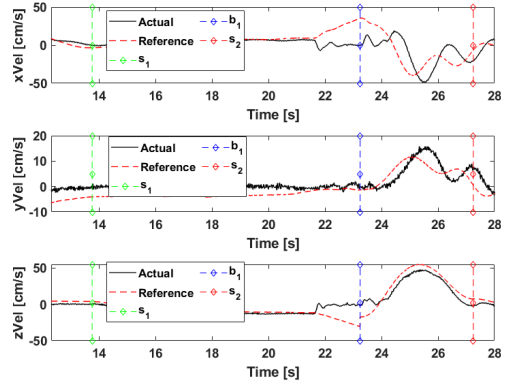}
  \caption{Actual Nailing Trajectory - Velocity Plot.}
  \label{fig:actual_traj_vel}
\end{figure}

\begin{figure}[htb]
  \centering
    \includegraphics[width=0.75\columnwidth,trim=30 185 44 185, clip]{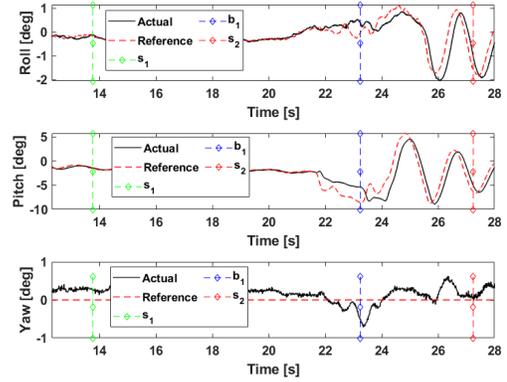}
  \caption{Actual Nailing Trajectory - Attitude Plot.}
  \label{fig:actual_traj_ori}
\end{figure}

From test set A, six experiments were performed on the same day in the same conditions (2 at 0$^\circ$, 2 at 15$^\circ$, 2 at 30$^\circ$). For each angle, a first test was made using the methods from Sec. \ref{sec:Planner} for guidance. After the test, nailing error was measured and the average vertical error was used to adjust the nailing setpoints for the second test. Nailing accuracy for set A is given in Figure \ref{fig:testA_box_plot}. Similar trends can be seen for test set A as were noticed for all tests. Additionally, the 2D nail points can be seen in Figure \ref{fig:shingle_plot_setA}. 
For 15$^\circ$ and 30$^\circ$ the correction allowed the second test to have all four nails above the shingle line and three below the tar line. 
Note that the nail points were slightly adjusted from Figure \ref{fig:shingle} to allow minor errors and still successfully nail the shingle. Violating the shingle line constraint or missing the shingle entirely would result in exposed nails and likely leaks. However, violating the tar line constraint occasionally would still produce a viable shingled roof, although it may be more susceptible to wind damage.  
Figures \ref{fig:actual_traj_pos}, \ref{fig:actual_traj_vel}, and \ref{fig:actual_traj_ori} show the position, velocity, and orientation data from the second test at $30^\circ$. A noticeable disturbance can be seen at about 22s when the vehicle contacts nail point 1 and the reference trajectory is no longer accurately followed. Also, note the small (2-3cm) error in x position immediately prior to nailing. This may account for some of the error at higher angles.



\section{Discussion}
\label{sec:Discussion}

Nailing accuracy results from Section \ref{sec:Results} offer evidence that our system can properly nail shingles to within a vertical 3cm gap. For all 12 nails at 30$^\circ$, our vertical range was only 2.9cm with a systematic bias of 6.4cm. Additionally, we have demonstrated a method to correct for this bias indicating that if the bias is properly accounted for and we maintain our precision we can accurately nail a shingle onto a roof at 30$^\circ$. Surprisingly, our range in vertical error was worse for 0$^\circ$ (5.5cm) and 15$^\circ$ (7.7cm) for all tests as well as test set A (3.6cm at 0$^\circ$, 6.1cm at 15$^\circ$)

The most likely source of vertical error is a systematic 2-3cm $x$ position error, combined with modelling errors. Due to the slope, any error in x position ($e_{x}$) will actually be amplified as $e_{x}$ / $cos(\alpha)$. The higher the slope the larger the effect. Additionally, we planned for a specific vehicle attitude during nailing. In the constant velocity trajectory, attitude is very near to hover, however, it isn't exact and could cause additional error.

Follow on work is required to build a complete autonomous aerial roofing system. Obviously, there won't be a motion capture system at a roofing site. However, this could be replaced by an onboard sensing system that recognizes shingle outlines on the approach trajectory, exactly where accurate state estimation is needed. Additionally, upgrading the system to a professional nailgun with an air hose and power tether would enable extended duration flights. The pneumatic tooltip would be easier to nail with so more aggressive trajectories may be attempted, as speed would be an important factor. Other aerial vehicles could carry shingles and other supplies up to the roof, where a ground-based robot or human with proper protection could lay down the shingles.

\section{Conclusion}
\label{sec:Conclusion}
This paper has presented the first experimental demonstration of an autonomous roofing system using an octocopter. The nailing problem was described and translated to an integrated octocopter-nailgun roofing platform. 
Waypoint-based guidance offering the required nailing contact force was demonstrated to show roofing success at different roof slopes. 
Future work includes extension to pneumatic air gun, power tether, and vision/depth based localization system.

\newpage
\bibliographystyle{IEEEtran}
\bibliography{IEEEabrv,refs}

\end{document}